\documentclass[english]{eptcs}
\usepackage[T1]{fontenc}
\usepackage[latin9]{inputenc}
\usepackage{float}
\usepackage{amsmath}
\usepackage{amssymb}
\usepackage{graphicx}

\makeatletter

\newcommand{\noun}[1]{\textsc{#1}}

\author{Benoit Desrochers,
\institute{DGA Tn, Lab-Sticc\\ Brest, France}
\email{benoit.desrochers@ensta-bretagne.org}
\and
Luc Jaulin
\institute{Ensta Bretagne, Lab-Sticc\\
Brest, France}
\email{luc.jaulin@gmail.com}
}
\def\titlerunning{Minkowski Operations for Localization}
\def\authorrunning{B. Desrochers, L. Jaulin}

\usepackage{babel}
\usepackage{stmaryrd}

\@ifundefined{showcaptionsetup}{}{%
 \PassOptionsToPackage{caption=false}{subfig}}
\usepackage{subfig}
\makeatother

\usepackage{babel}
\begin{document}

\title{Minkowski Operations of Sets \\
with Application to Robot Localization}
\maketitle
\begin{abstract}
This papers shows that using separators, which is a pair of two complementary
contractors, we can easily and efficiently solve the localization
problem of a robot with sonar measurements in an unstructured environment.
We introduce separators associated with the Minkowski sum and the
Minkowski difference in order to facilitate the resolution. A test-case
is given in order to illustrate the principle of the approach. 
\end{abstract}

\section{Introduction}

Interval analysis \cite{Moore66} is a tool which makes it possible
to compute with sets even when nonlinear functions are involved \cite{Kearfott96}
in the definition of the sets. Interval methods are generally used
to solve equations or optimization problems \cite{Hansen92} but can
also been used to solve set-membership problems where the sets are
represented by subpavings \cite{JaulinBook01}. The efficiency of
interval algorithms can be improved by the use of contractors \cite{ChabertJaulin09}
or (separators \cite{DesrochersEAAI2014} which correspond to pairs
of contractors). 

This paper deals with localization of a robot with sonar rangefinders
in a unstructured environment. This problem is considered as difficult
due to the fact that the sonar returns a measurement under the form
of an impact point inside an emission cone. This specific type of
measurement makes the problem partially observable. Moreover, our
environment is not represented by geometric features such as segments
or disks, but by an image which cannot be translated into equations.
Now, as shown by Sliwka \cite{Sliwka:elpaso:11}, an unstructured
map can be cast into a contractor form which allows us to use contractor/separator
algebra.

Here, we propose first to use a separator-based method to perform
a reliable simulation necessary to generate realistic data (see, \emph{e.g.},
\cite{taha:15:acumen} for a survey on reliable simulation). Then,
once these data have been generated, we consider the inverse problem,
\emph{i.e.}, the robot localization with large-cone sonar measurements
in an unstructured map. This problem has never been considered yet,
to our knowledge at least in an unstructured environment (see \emph{e.g.},
\cite{JaulinCHI99,KiefferMISC99,Jaulin02Robab,Leveque97,colle_galerne13}
in the case where the map is made with geometrical features). We will
also show the link with Minkowski operations and propose separator
counterparts for these operations. 

Section \ref{sec:Contractors-and-Separators} recalls the basic notions
on contractors and separators needed to understand our approach. Section
\ref{sec:Set-to-set-Transformation} presents the concept of set-to-set
transform and shows how our localization problem can be solved with
separators. Section \ref{sec:Minkowski-sum-and} proposes to formulate
the Minkowski operations as a specific set-to-set transform, corresponding
to translations. Section \ref{sec:Localization-in-an} illustrates
the application of the Minkowski operation to the problem of localization
of a robot in an unstructured environment.  Section \ref{sec:Conclusion}
concludes the paper.

\section{Contractors and Separators} \label{sec:Contractors-and-Separators}

This section recalls the basic notions on intervals, contractors and
separators that are needed to understand the contribution of this
paper. An \emph{interval} of $\mathbb{R}$ is a closed connected set
of $\mathbb{R}$. A box $\left[\mathbf{x}\right]$ of $\mathbb{R}^{n}$
is the Cartesian product of $n$ intervals. 

A \emph{contractor} $\mathcal{C}$ is an operator $\mathbb{IR}^{n}\mapsto\mathbb{IR}^{n}$
such that 
\begin{equation}
\begin{array}{lll}
\mathcal{C}([\mathbf{x}])\subset[\mathbf{x}] &  & \text{(contractance)}\\{}
[\mathbf{x}]\subset\left[\mathbf{y}\right]\text{ }\Rightarrow\text{ }\mathcal{C}([\mathbf{x}])\subset\mathcal{C}([\mathbf{y}]). &  & \text{(monotonicity)}
\end{array}
\end{equation}

\begin{figure}
\centering{}\includegraphics[width=0.45\textwidth]{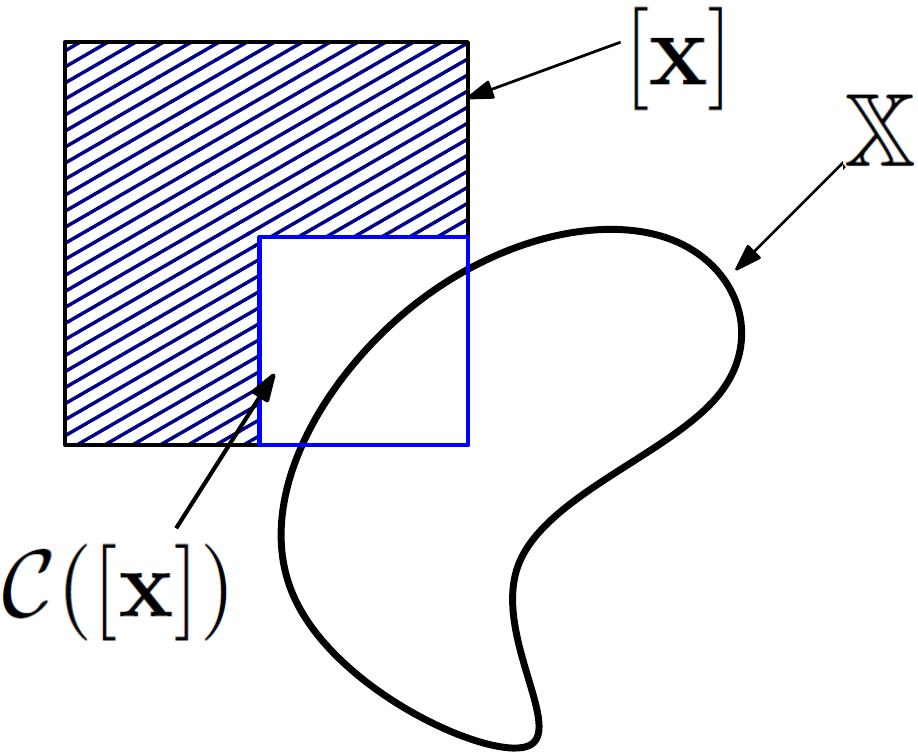}\caption{Contractor consistent with to the set $\mathbb{X}$}
\label{fig:contractorX}
\end{figure}

We define the inclusion between two contractors $\mathcal{C}_{1}$
and $\mathcal{C}_{2}$ as follows: 
\begin{equation}
\mathcal{C}_{1}\subset\mathcal{C}_{2}\Leftrightarrow\forall\left[\mathbf{x}\right]\in\mathbb{IR}^{n}\text{, }\mathcal{C}_{1}([\mathbf{x}])\subset\mathcal{C}_{2}([\mathbf{x}]).
\end{equation}
A set $\mathbb{X}$ is \emph{consistent} (See Figure \ref{fig:contractorX})
with the contractor $\mathcal{C}$ (we will write $\mathbb{X}\sim\mathcal{C}$)
if for all $\left[\mathbf{x}\right]$, we have 
\begin{equation}
\mathcal{C}([\mathbf{x}])\cap\mathbb{X}=[\mathbf{x}]\cap\mathbb{X}.\label{eq:ctc_consistent}
\end{equation}

\noindent Two contractors $\mathcal{C}$ and $\mathcal{C}_{1}$ are
equivalent (we will write $\mathcal{C}\sim\mathcal{C}_{1}$) if we
have: 
\begin{equation}
\mathbb{X}\sim\mathcal{C}\Leftrightarrow\mathbb{X}\sim\mathcal{C}_{1}.
\end{equation}
A contractor $\mathcal{C}$ is \emph{minimal} if for any other contractor
$\mathcal{C}_{1}$, we have the following implication 
\begin{equation}
\mathcal{C}\sim\mathcal{C}_{1}\Rightarrow\mathcal{C}\subset\mathcal{C}_{1}.
\end{equation}

\begin{figure}
\centering%
\begin{minipage}[t]{0.45\linewidth}%
\includegraphics[width=1\linewidth]{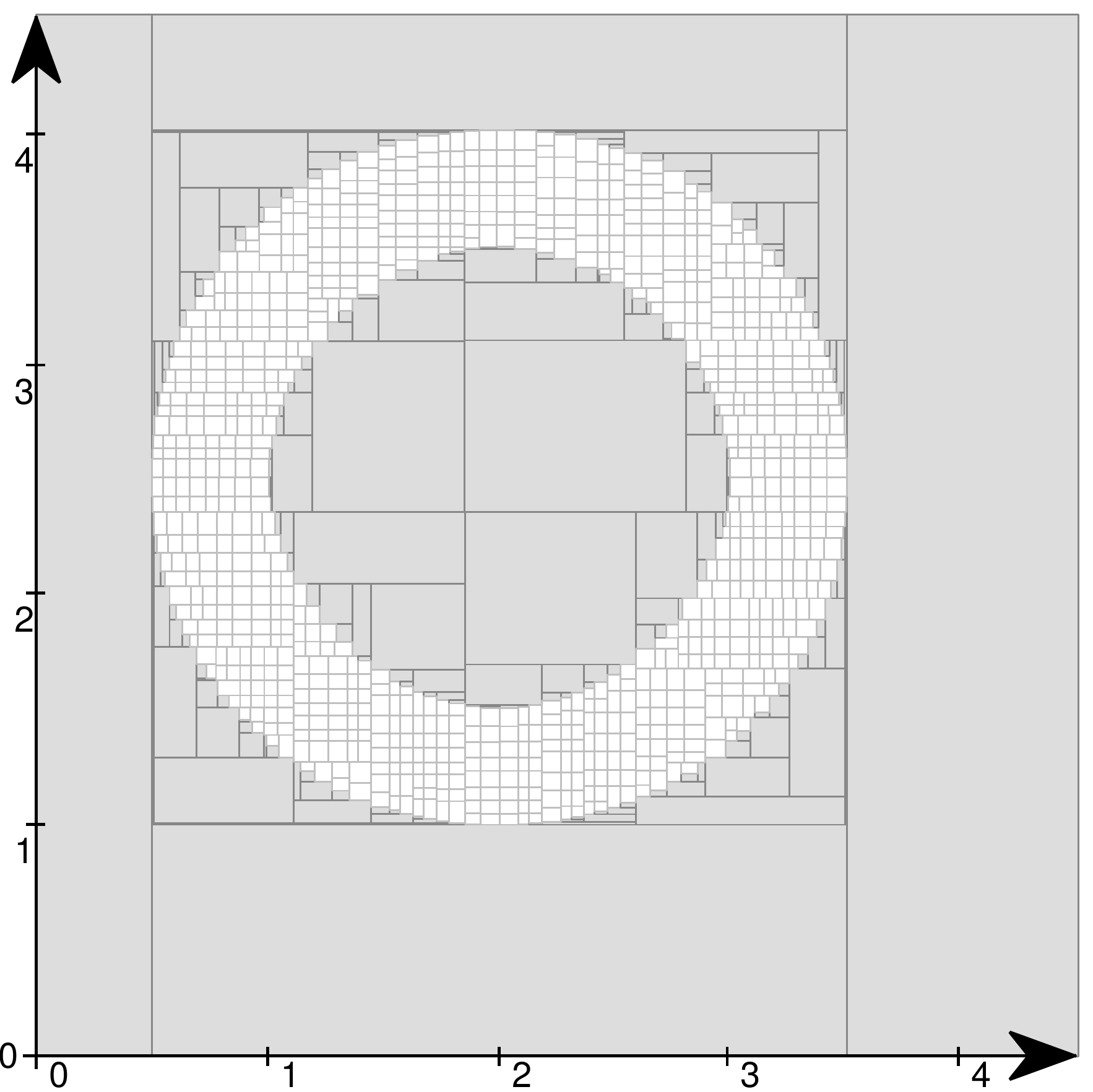}%
\end{minipage}\hspace*{\fill}%
\begin{minipage}[t]{0.45\columnwidth}%
\begin{center}
\includegraphics[width=1\linewidth]{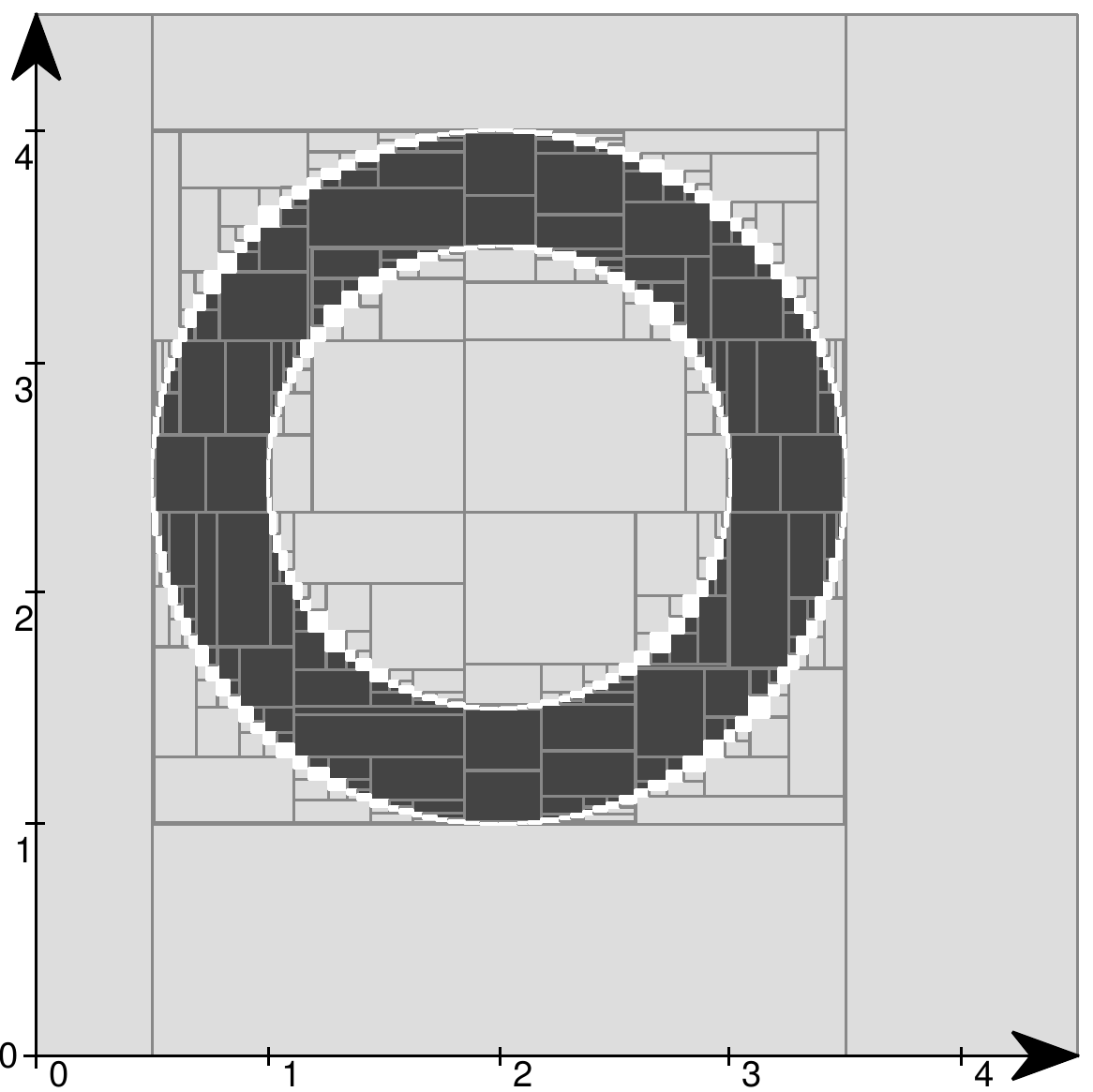}
\par\end{center}%
\end{minipage} \caption{Paving associated to Example 1, Left: paving obtained using the contractor,
Right: paving obtained using the separator. Dark gray boxes belong
$\mathbb{X}$ (the ring); light gray boxes are outside $\mathbb{X}$.
No conclusion can be given on the white boxes. }
\label{fig:OuterApproximation}
\end{figure}

\noindent \textbf{Example 1}. The minimal contractor\textbf{ $\mathcal{C}_{\mathbb{X}}$
}consistent with the set 
\begin{equation}
\mathbb{X}=\left\{ \mathbf{x}\in\mathbb{R}^{2},(x_{1}-2)^{2}+(x_{2}-2.5)^{2}\in\left[1,4\right]\right\} 
\end{equation}
can be built using a forward-backward constraint propagation\textbf{
\cite{Ben99} \cite{Drevelle:Bonnifait:12}}. The contractor \textbf{$\mathcal{C}_{\mathbb{X}}$
}can be used by a paver to obtain an outer approximation for $\mathbb{X}$.
This is illustrated by Figure\textbf{~\ref{fig:OuterApproximation}}
(left) where\textbf{ $\mathcal{C}_{\mathbb{X}}$} removes parts of
the space outside\textbf{ $\mathbb{X}$ }(painted light-gray). But
due to the consistency property (see~Equation \eqref{eq:ctc_consistent})\textbf{
$\mathcal{C}_{\mathbb{X}}$ }has no effect on boxes included in\textbf{
}$\mathbb{X}$. A box partially included in \textbf{$\mathbb{X}$
}can not be eliminated and is bisected, except if its length is larger
than an given value $\varepsilon$. The contractor\textbf{ $\mathcal{C}_{\mathbb{X}}$
}only provides an outer approximation of \textbf{$\mathbb{X}$}. 

\noindent If\textbf{ $\mathcal{C}_{1}$} and \textbf{$\mathcal{C}_{2}$
}are two contractors, we define the following operations\textbf{~\cite{ChabertJaulin09}.
}
\begin{eqnarray}
(\mathcal{C}_{1}\cap\mathcal{C}_{2})([\mathbf{x}]) & = & \mathcal{C}_{1}([\mathbf{x}])\cap\mathcal{C}_{2}([\mathbf{x}])\\
(\mathcal{C}_{1}\sqcup\mathcal{C}_{2})([\mathbf{x}]) & = & \mathcal{C}_{1}([\mathbf{x}])\sqcup\mathcal{C}_{2}([\mathbf{x}])\\
(\mathcal{C}_{1}\circ\mathcal{C}_{2})([\mathbf{x}]) & = & \mathcal{C}_{1}\left(\mathcal{C}_{2}([\mathbf{x}])\right)
\end{eqnarray}
where $\sqcup$ is the \emph{union hull} defined by
\begin{equation}
[\mathbf{x}]\sqcup[\mathbf{y}]=\left[[\mathbf{x}]\cup[\mathbf{y}]\right].
\end{equation}

\begin{figure}
\centering{}\includegraphics[width=0.45\linewidth]{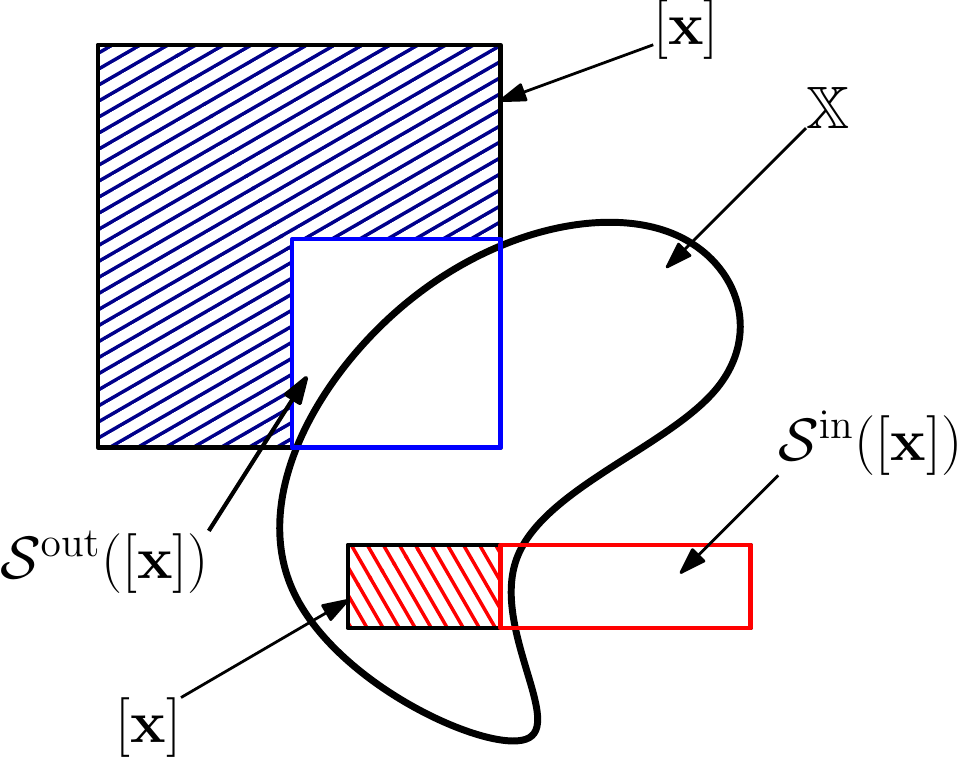}
\caption{Illustration of a separator on two different initial boxes. The outer
contractor removes the blue dashed area and the red dashed area is
removed by the inner contractor}
\label{fig:InnerOuterContractor} 
\end{figure}

In order to characterize an inner and outer approximation of the solution
set, we introduce the notion of \emph{separator}.\\
 A \emph{separator} $\mathcal{S}$ is a pair of contractors $\left\{ \mathcal{S}^{\text{in}},\mathcal{S}^{\text{out}}\right\} $
such that, for all $\left[\mathbf{x}\right]\in\mathbb{IR}^{n}$, we
have 
\begin{equation}
\begin{array}{lll}
\mathcal{S}^{\text{in}}(\left[\mathbf{x}\right])\cup\mathcal{S}^{\text{out}}([\mathbf{x}])=\left[\mathbf{x}\right] &  & \text{(complementarity).}\end{array}\text{ }\label{eq:sep:complementarity}
\end{equation}
A set $\mathbb{X}$ is \emph{consistent} with the separator $\mathcal{S}$
(we will write $\mathbb{X}\sim\mathcal{S}$), if 
\begin{equation}
\mathbb{X}\sim\mathcal{S}^{\text{out}}\text{ and }\overline{\mathbb{X}}\sim\mathcal{S}^{\text{in}}\text{,}\label{eq:set:sep:consistent}
\end{equation}
where $\overline{\mathbb{X}}=\{\mathbf{x}\ |\ \mathbf{x}\notin\mathbb{X}\}$.
This notion of separator is illustrated by Figure~\ref{fig:InnerOuterContractor}.\\
We define the inclusion between two separators $\mathcal{S}_{1}$
and $\mathcal{S}_{2}$ as follows 
\begin{equation}
\mathcal{S}_{1}\subset\mathcal{S}_{2}\Leftrightarrow\mathcal{S}_{1}^{\text{in}}\subset\mathcal{S}_{2}^{\text{in}}\text{ and }\mathcal{S}_{1}^{\text{out}}\subset\mathcal{S}_{2}^{\text{out}}.
\end{equation}
A separator $\mathcal{S}$ is \emph{minimal} if 
\begin{equation}
\mathcal{S}_{1}\subset\mathcal{S}\Rightarrow\mathcal{S}_{1}=\mathcal{S}.
\end{equation}
It is trivial to check that $\mathcal{S}$ is minimal implies that
the two contractors $\mathcal{S}^{\text{in}}$ and $\mathcal{S}^{\text{out}}$
are both minimal. If we define the following operations 
\begin{equation}
\begin{array}{cccc}
\mathcal{S}_{1}\cap\mathcal{S}_{2} & = & \left\{ \mathcal{S}_{1}^{\text{in}}\cup\mathcal{S}_{2}^{\text{in}},\mathcal{S}_{1}^{\text{out}}\cap\mathcal{S}_{2}^{\text{out}}\right\}  & \text{(intersection)}\\
\mathcal{S}_{1}\cup\mathcal{S}_{2} & = & \left\{ \mathcal{S}_{1}^{\text{in}}\cap\mathcal{S}_{2}^{\text{in}},\mathcal{S}_{1}^{\text{out}}\cup\mathcal{S}_{2}^{\text{out}}\right\}  & \text{(union)}
\end{array}\label{eq:sep:nary:op}
\end{equation}
then we have \cite{DesrochersEAAI2014}
\begin{equation}
\left\{ \begin{array}{ccc}
\mathcal{S}_{1} & \sim & \mathbb{X}_{1}\\
\mathcal{S}_{2} & \sim & \mathbb{X}_{2}
\end{array}\right.\Rightarrow\left\{ \begin{array}{ccc}
\mathcal{S}_{1}\cap\mathcal{S}_{2} & \sim & \mathbb{X}_{1}\cap\mathbb{X}_{2}\\
\mathcal{S}_{1}\cup\mathcal{S}_{2} & \sim & \mathbb{X}_{1}\cup\mathbb{X}_{2}
\end{array}\right.
\end{equation}
Other operations on separators such as the complement or the projection
can also be considered \cite{DesrochersEAAI2014}. 

\textbf{Example 2}. Consider the set\textbf{ $\mathbb{X}$} of Example~1.
From the contractor consistent with 
\begin{equation}
\mathbb{\overline{X}}=\left\{ \mathbf{x}\in\mathbb{R}^{2},(x_{1}-2)^{2}+(x_{2}-2.5)^{2}\notin\left[1,4\right]\right\} ,
\end{equation}
we can build a separator\textbf{ $\mathcal{S}_{\mathbb{X}}$ }for
$\mathbb{X}$. An inner and outer approximation of $\mathbb{X}$ obtained
by a paver based on \textbf{$\mathcal{S}_{\mathbb{X}}$ }is depicted
on Figure~\ref{fig:OuterApproximation}. The dark gray area is inside
$\mathbb{X}$ and light gray is outside. The minimality property of
the separators can be observed by the fact that all contracted boxes
of the subpaving touch the boundary of $\mathbb{X}$. Therefore, we
are now able to quantify the pessimism introduced by the paver. 

\section{Set-to-set Transform \label{sec:Set-to-set-Transformation}}

\textbf{Notation}. Consider a function
\begin{equation}
\mathbf{f}:\left\{ \begin{array}{ccc}
\mathbb{R}^{n}\times\mathbb{R}^{p} & \longrightarrow & \mathbb{R}^{m}\\
(\mathbf{a},\mathbf{p}) & \longrightarrow & \mathbf{f}(\mathbf{a},\mathbf{p})
\end{array}\right.
\end{equation}
 For a given $\mathbf{p}\in\mathbb{R}^{p}$, $\mathbb{A}\subset\mathbb{R}^{n}$,
$\mathbb{B}\subset\mathbb{R}^{m}$, $\mathbb{Z}\subset\mathbb{R}^{n}\times\mathbb{R}^{p}$
, we shall use the following notations:
\begin{equation}
\begin{array}{cl}
\mathbf{f}(\mathbb{A},\mathbf{p}) & =\left\{ \mathbf{b}|\exists\mathbf{a}\in\mathbb{A},\,\mathbf{b}=\mathbf{f}(\mathbf{a},\mathbf{p})\right\} \\
\mathbf{f}^{-1}(\mathbb{B}) & =\left\{ \mathbf{z}=(\mathbf{a},\mathbf{p})\,|\exists\mathbf{b}\in\mathbb{B},\,\mathbf{b}=\mathbf{f}(\mathbf{a},\mathbf{p})\right\} \\
proj_{\mathbf{p}}(\mathbb{Z}) & =\left\{ \mathbf{p}\,|\exists\mathbf{a},\,(\mathbf{a},\mathbf{p})\in\mathbb{Z}\right\} \\
\overline{\mathbb{Z}} & =\left\{ \mathbf{z}\,|\,\mathbf{z}\notin\mathbb{Z}\right\} 
\end{array}
\end{equation}
 Many sets that are defined with quantifiers can be defined in terms
of projection, inversion, complement and composition. 

An important problem where these operations occur is the set-to-set
transform which is now defined.

\textbf{Set-to-set transform}. Consider the set defined by :
\begin{equation}
\mathbb{P}=\{\mathbf{p}\in\mathbb{R}^{p}\mid\mathbf{f}(\mathbb{A},\mathbf{p})\subset\mathbb{B}\}.\label{eq:fAsubsetB}
\end{equation}
The vector \textbf{$\mathbf{p}$} corresponds to a parameter vector
associated to a transformation. A transformation $\mathbf{p}$ is
consistent, if after transformation of $\mathbb{A}$, the set $\mathbb{A}$
is included inside $\mathbb{B}$. We have
\begin{equation}
\begin{array}{cl}
 & \mathbf{f}(\mathbb{A},\mathbf{p})\subset\mathbb{B}\\
\Leftrightarrow & \forall\mathbf{a}\in\mathbb{A},\mathbf{f}(\mathbf{a},\mathbf{p})\in\mathbb{B}\\
\Leftrightarrow & \neg\exists\mathbf{a}\in\mathbb{A},\mathbf{f}(\mathbf{a},\mathbf{p})\in\overline{\mathbb{B}}\\
\Leftrightarrow & \neg\exists\mathbf{a}\in\mathbb{A},(\mathbf{a},\mathbf{p})\in\mathbf{f}^{-1}\left(\overline{\mathbb{B}}\right)\\
\Leftrightarrow & \neg\exists\mathbf{a},(\mathbf{a},\mathbf{p})\in\mathbb{A}\times\mathbb{R}^{p}\,\wedge\,(\mathbf{a},\mathbf{p})\in\mathbf{f}^{-1}\left(\overline{\mathbb{B}}\right).
\end{array}
\end{equation}
As a consequence
\begin{equation}
\mathbb{P}=\overline{proj_{\mathbf{p}}\{\left(\mathbb{A}\times\mathbb{R}^{p}\right)\cap\mathbf{f}^{-1}\left(\overline{\mathbb{B}}\right)\}}.
\end{equation}

Therefore, if we have separators $\mathcal{S}_{\mathbb{A}},\mathcal{S}_{\mathbb{B}}$
for $\mathbb{A},\mathbb{B}$ then a separator $\mathcal{S}_{\mathbb{P}}$
for $\mathbb{P}$ can be obtained using the separator algebra \cite{DesrochersEAAI2014}.
It is given by
\begin{equation}
\mathcal{S}_{\mathbb{P}}=\overline{proj_{\mathbf{p}}\{\left(\mathbb{\mathcal{S}_{\mathbb{A}}}\times\mathcal{S}_{\mathbb{R}^{p}}\right)\cap\mathbf{f}^{-1}\left(\overline{\mathcal{S}_{\mathbb{B}}}\right)\}}.
\end{equation}
Combining this separator with a paver, we are able to obtain an inner
and outer approximation of $\mathbb{P}$.

\begin{figure}[H]
\begin{centering}
\includegraphics[width=0.4\linewidth]{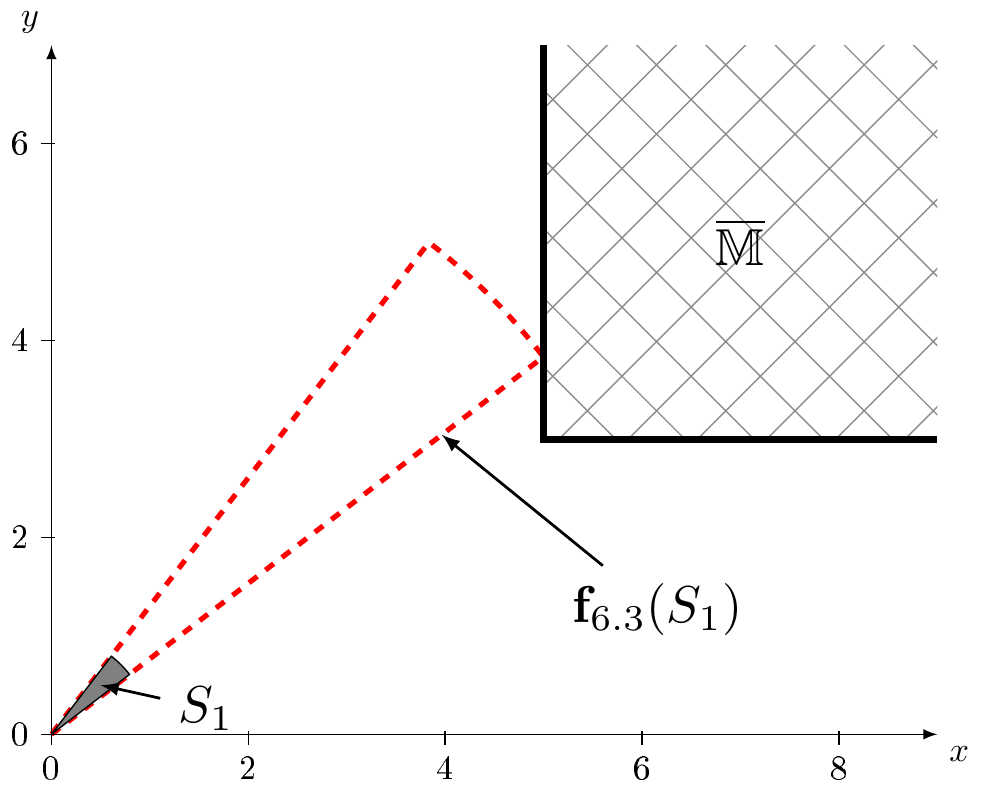}
\par\end{centering}
\caption{The map $\mathbb{M}$ is represented by the white space outside the
hatched area while the unit pie $\mathbb{S}_{1}$ is painted in gray.
The red pie represents the lower upper bound of $\mathbb{\ensuremath{D}}$
which almost touches the border of the map. \label{fig:Example1}}
\end{figure}

\textbf{Example 1}: A robot at position (0,0) in inside an environment
defined by the map
\begin{equation}
\mathbb{M}=\{\mathbf{x}\in\mathbb{R}^{2}\mid x_{1}<5\text{ or }x_{2}<3\}.
\end{equation}
It emits an ultrasonic sound in the cone with angles $\frac{\pi}{4}\pm\frac{\pi}{24}$.
For a simulation purpose, we want to compute the distance returned
by the sonar. This distance corresponds to the shortest distance inside
the emission cone to the complementary of the map: 
\[
d=\inf\{d\mid\mathbf{f}(\mathbb{S}_{1},d)\cap\overline{\mathbb{M}}\neq\emptyset\}
\]
or equivalently
\begin{equation}
d=\sup\{d\mid\mathbf{f}(\mathbb{S}_{1},d)\subset\mathbb{M}\},
\end{equation}
 where $\mathbb{S}_{1}$ is the unit cone defined by 

\begin{equation}
\mathbb{S}_{1}=\{(x,y)\mid x^{2}+y^{2}<1\text{ and }atan2(y,x)\in[\frac{5\pi}{24},\frac{7\pi}{24}]\}.
\end{equation}
and $\mathbf{f}(\mathbf{x},d)=d\cdot\mathbf{x}$ is the scaling function.
To solve our problem, we first characterize the set:

\begin{equation}
\mathbb{D}=\{d\mid\mathbf{f}(\mathbb{S}_{1},d)\subset\mathbb{M}\}
\end{equation}
which corresponds to a set-to-set transform problem and we get
\begin{equation}
[0,6.2988]\subset\mathbb{D}\subset[0,6.3085].
\end{equation}
The situation is depicted on Figure \ref{fig:Example1}. As a consequence,
the true distance $d$ returned by the sensor satisfies $d\in[6.2988,6.3085]$

\section{Minkowski sum and difference} \label{sec:Minkowski-sum-and}

Minkowski operations are used in morphological mathematics to perform
dilation or inflation of sets. As it will be shown in Section 4, it
can also be used for localization. Efficient algorithms (see e.g.,
\cite{Shoji91}) have been proposed to perform Minkowski operations
with sets represented by subpavings. In this section, we show Minkowski
sum and difference can be see as a set-to-set transform. This will
allow us to build separators for these Minkowski operations. 

\subsection{Minkowski difference}

Given two sets $\mathbb{A}\subset\mathcal{P}(\mathbb{R}^{n})$, $\mathbb{B}\subset\mathcal{P}(\mathbb{R}^{n})$,
the Minkowski difference \cite{Najman2010}, denoted $\ominus$, defined
by
\begin{align}
\mathbb{B}\ominus\mathbb{A} & =\{\mathbf{p}\mid\mathbb{A}+\mathbf{p}\subset\mathbb{B}\}.\label{eq:BminusA}
\end{align}
\textbf{Proposition 1}. Given two separator $\mathcal{S}_{\mathbb{A}}$
and $\mathcal{S}_{\mathbb{B}}$ for $\mathbb{A}$ and \noun{$\mathbb{B}$}.
Define the Minkowski difference of two separators as
\begin{equation}
\mathcal{S}_{\mathbb{B}}\ominus\mathcal{S}_{\mathbb{A}}=\overline{proj_{\mathbf{p}}\{\left(\mathbb{\mathcal{S}_{\mathbb{A}}}\times\mathcal{S}_{\mathbb{R}^{n}}\right)\cap\mathbf{f}^{-1}\left(\overline{\mathcal{S}_{\mathbb{B}}}\right)\}}
\end{equation}
where $\mathbf{f}(\mathbf{p},\mathbf{a})=\mathbf{a}+\mathbf{p}$.
The operator $\mathcal{S}_{\mathbb{B}}\ominus\mathcal{S}_{\mathbb{A}}$is
a separator for $\mathbb{B}\ominus\mathbb{A}$.

\textbf{Proof}. Computing the Minkowski difference can be seen as
a specific set-to-set transform problem where $\mathbf{f}(\mathbf{p},\mathbf{a})=\mathbf{a}+\mathbf{p}$
, \emph{i.e.}, the transformation corresponds to a translation of
vector\texttt{ }$\mathbf{p}$. As a consequence
\begin{equation}
\mathbb{B}\ominus\mathbb{A}=\overline{proj_{\mathbf{p}}\{\left(\mathbb{A}\times\mathbb{R}^{p}\right)\cap\mathbf{f}^{-1}\left(\overline{\mathbb{B}}\right)\}}.\blacksquare
\end{equation}
A separator can thus be built for $\mathbb{B}\ominus\mathbb{A}$ and
a paver is then able to characterize $\mathbb{B}\ominus\mathbb{A}.$

\begin{figure}[h]
\centering{}\includegraphics[width=0.33\linewidth]{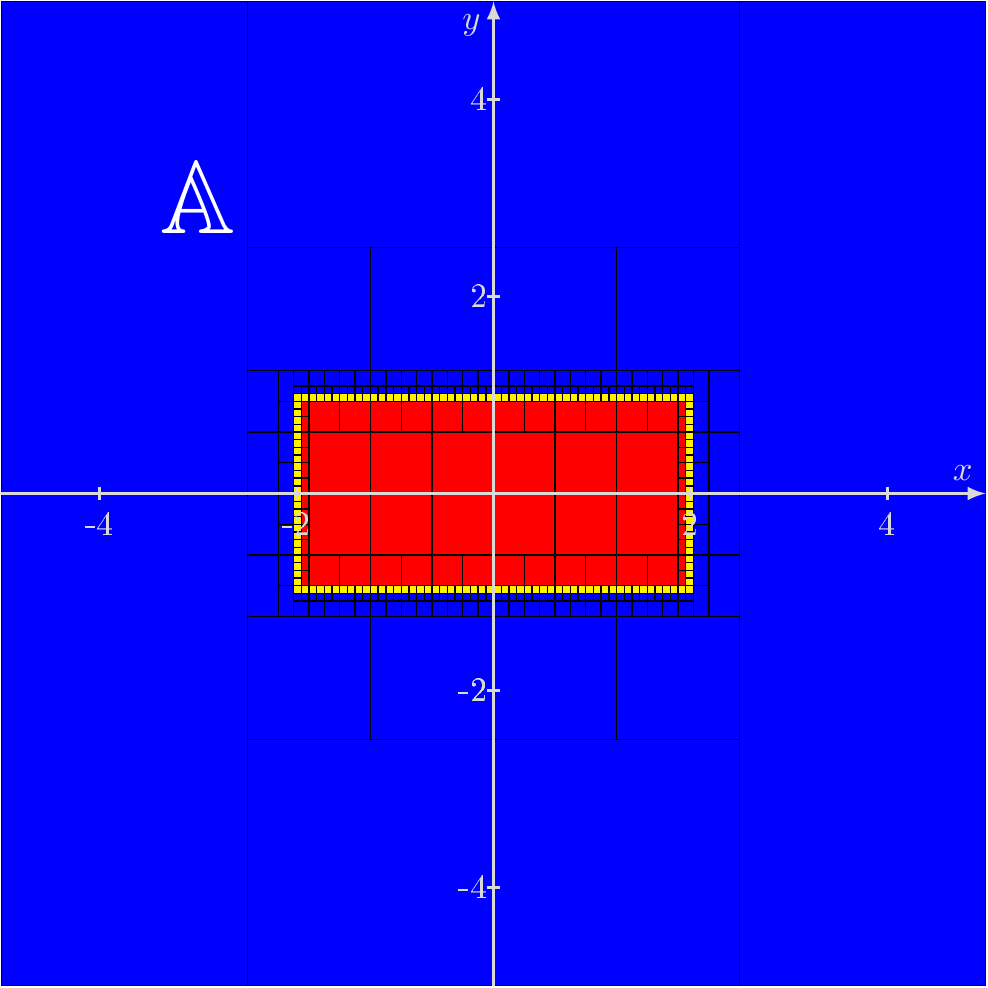}\hfill{}\includegraphics[width=0.33\linewidth]{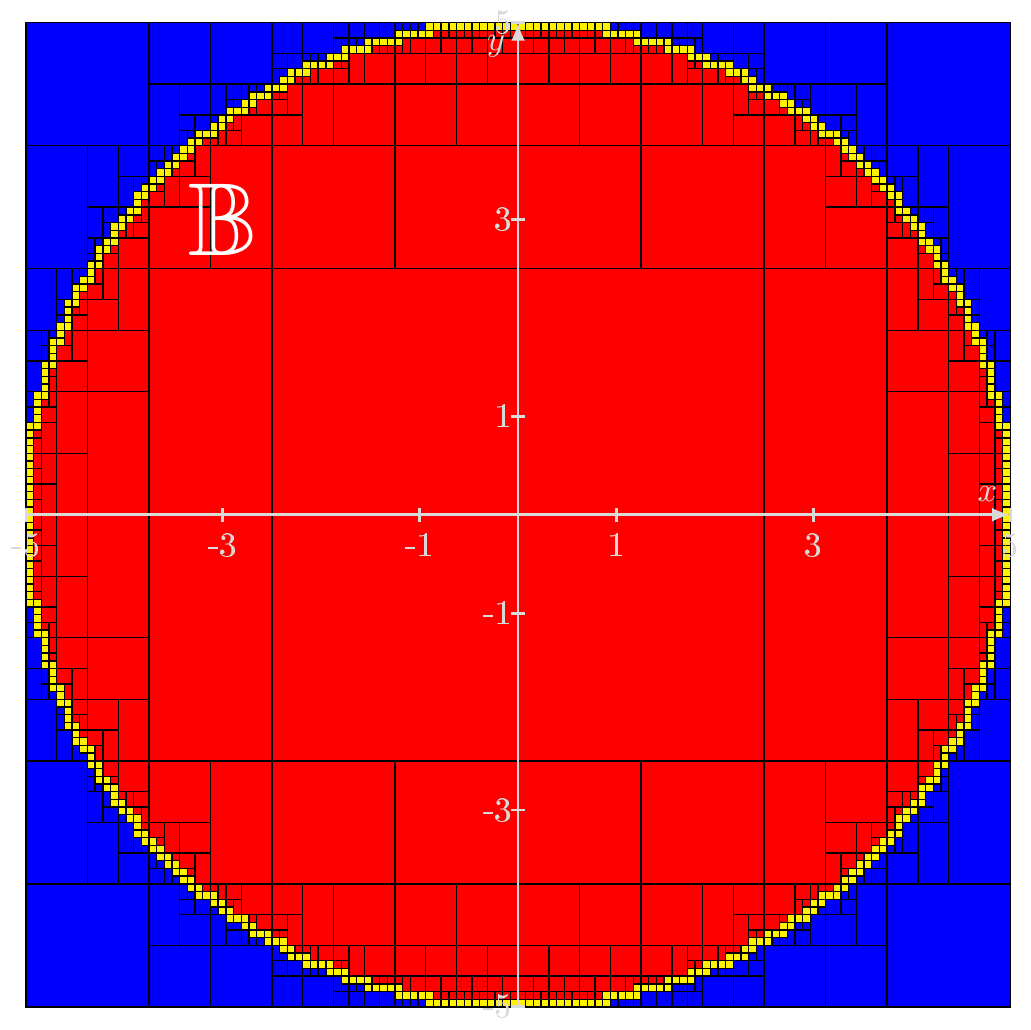}\hspace*{\fill}\includegraphics[width=0.33\linewidth]{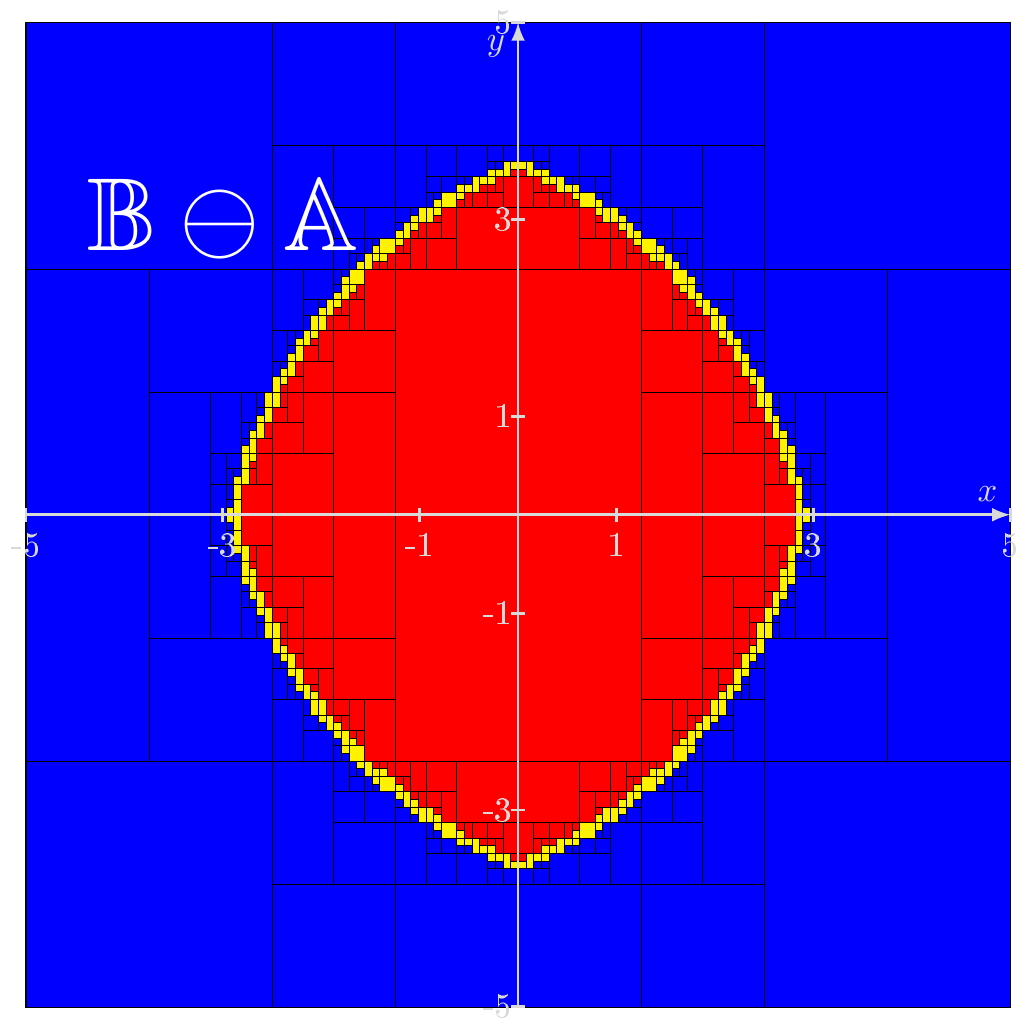}\caption{Minkowski difference of the disk $\mathbb{\ensuremath{B}}$ and the
rectangle $\mathbb{A}$ \label{fig:SetTranslation}}
\end{figure}

\textbf{Example} \textbf{2}: Let $\mathbb{A}$ be a rectangle of side's
length of 4 x 2, and $\mathbb{B}$ be a disk of radius 5. The resulting
solution set $\mathbb{B}\ominus\mathbb{A}$ is depicted in Figure
\ref{fig:SetTranslation}.

\subsection{Minkowski addition}

Given two sets $\mathbb{A}\in\mathcal{P}(\mathbb{R}^{n})$, $\mathbb{B}\in\mathcal{P}(\mathbb{R}^{n})$,
the Minkowski sum, denoted by $\oplus$, is defined by:
\begin{align}
\mathbb{A}\oplus\mathbb{B} & =\{\mathbf{a}+\mathbf{b},\mathbf{a}\in\mathbb{A},\mathbf{b}\in\mathbb{B}\}.
\end{align}

\textbf{Proposition 2}. Given two separators $\mathcal{S}_{\mathbb{A}}$
and $\mathcal{S}_{\mathbb{B}}$ for $\mathbb{A}$ and \noun{$\mathbb{B}$}.
The Minkowski sum of two separators defined by
\begin{equation}
\mathcal{S}_{\mathbb{A}}\oplus\mathcal{S}_{\mathbb{B}}=\overline{\overline{\mathbb{\mathcal{S}_{\mathbb{B}}}}\ominus-\mathbb{\mathcal{S}_{\mathbb{A}}}}
\end{equation}
 is a separator for $\mathbb{A}\oplus\mathbb{B}$.

\textbf{Proof}. We have:
\begin{equation}
\begin{array}{ccl}
\mathbb{A}\oplus\mathbb{B} & = & \{\mathbf{p}\mid\exists\mathbf{a}\in\mathbb{A},\exists\mathbf{b}\in\mathbb{B},\mathbf{p}=\mathbf{a}+\mathbf{b}\}\\
 & = & \{\mathbf{p}\mid\exists\mathbf{a}\in\mathbb{A},\exists\mathbf{b}\in\mathbb{B},\mathbf{p}-\mathbf{a}=\mathbf{b}\}\\
 & = & \{\mathbf{p}\mid(\mathbf{p}-\mathbb{A})\cap\mathbb{B}\neq\emptyset\}\\
 & = & \overline{\{\mathbf{p}\mid(\mathbf{p}-\mathbb{A})\cap\mathbb{B}=\emptyset\}}\\
 & = & \overline{\{\mathbf{p}\mid(\mathbf{p}+(-\mathbb{A}))\subset\overline{\mathbb{B}}\}}\text{(see Eq \ref{eq:BminusA})}\\
 & = & \overline{\overline{\mathbb{B}}\ominus-\mathbb{A}}.
\end{array}
\end{equation}
Thus, a separator for the set $\mathbb{A}\oplus\mathbb{B}$ is $\overline{\overline{\mathbb{\mathcal{S}_{\mathbb{B}}}}\ominus-\mathbb{\mathcal{S}_{\mathbb{A}}}}$.$\blacksquare$

\begin{figure}[H]
\hspace*{\fill}\includegraphics[width=0.4\linewidth]{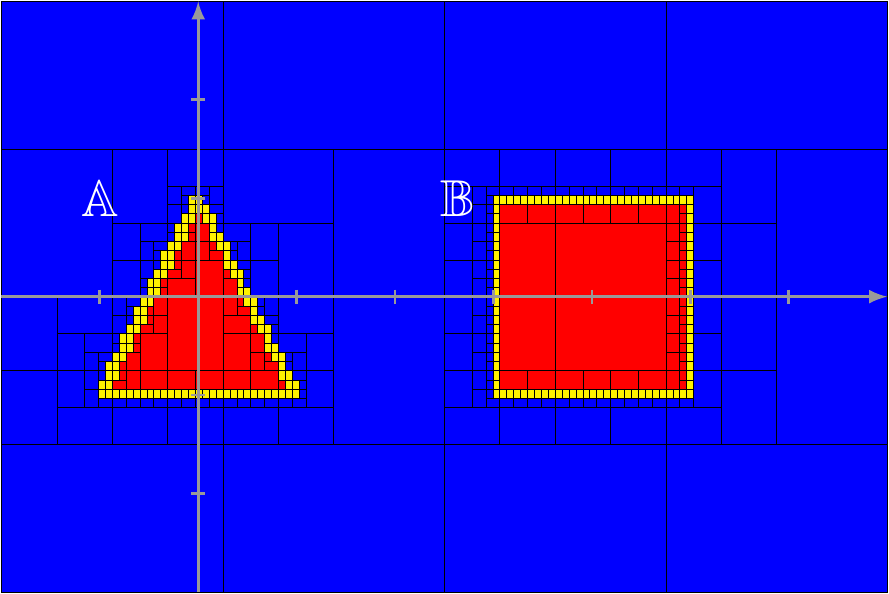}\hspace*{\fill}\includegraphics[width=0.4\linewidth]{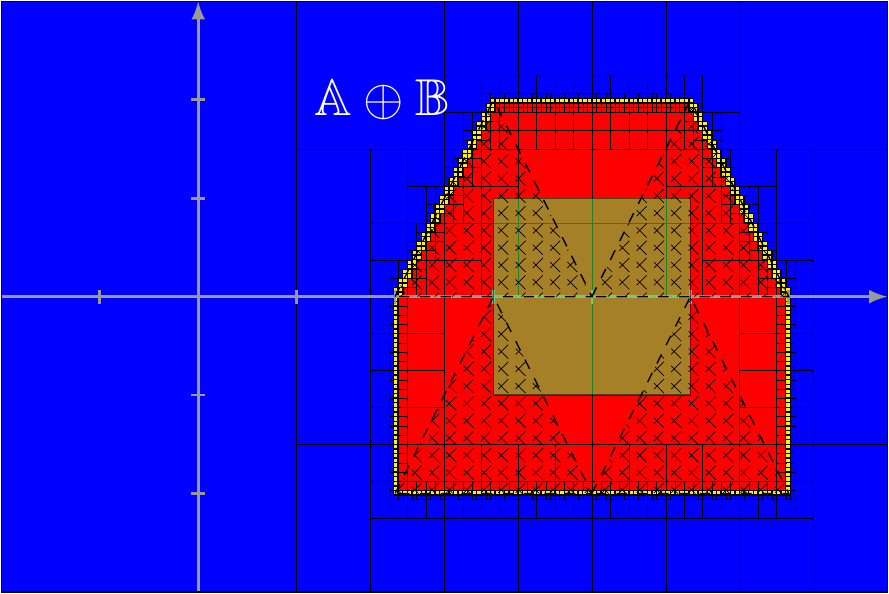}\hspace*{\fill}

\caption{Minkowski sum of a square $\mathbb{B}$ and a triangle $\mathbb{A}$
computed using set membership algorithm.}

\label{fig:AplusB}
\end{figure}

\textbf{Example 3:} Consider a triangle $\mathbb{A}$ and a square
$\mathbb{B}$. The Minkowski addition $\mathbb{A}\text{\ensuremath{\oplus\mathbb{B}}}$
is shown on Figure \ref{fig:AplusB}. 

\section{Localization in an unstructured environment\label{sec:Localization-in-an}}

Consider a robot $\mathcal{R}$ at position $\mathbf{p}=(p_{1},p_{2})$
in an unstructured environment described by the set $\mathbb{M}$.
We assume that the heading $\theta$ of $\mathcal{R}$ is known with
a good accuracy (for instance, by using a compass) and doesn't need
to be estimated. The robot is equipped several sonars which return
the distance between the robot and the map with respect to the emission
cone of the sonar. This section deals with the localization of the
robots using the set-to-set transform. Several authors have already
studied this problem using interval analysis \cite{Leveque97,meizel02,colle_galerne13}
but in an environment made with segments. 

\begin{figure}[H]
\subfloat[Emission cone.]{\begin{centering}
\includegraphics[width=0.45\linewidth]{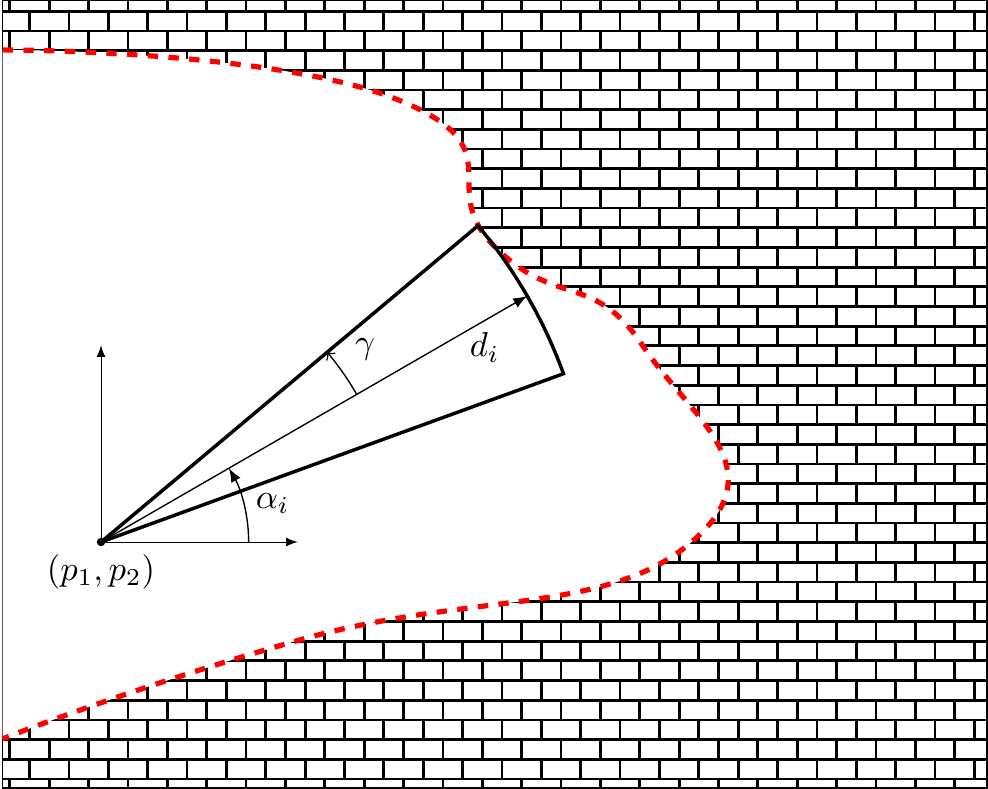}
\par\end{centering}
\label{fig:sensorModel}}\hspace*{\fill}\subfloat[The free sector $\mathbb{S}_{i}$ is represented by the dotted area
while the gray pie $\Delta\mathbb{S}_{i}$ contains the impact point.]{\begin{centering}
\includegraphics[width=0.45\linewidth]{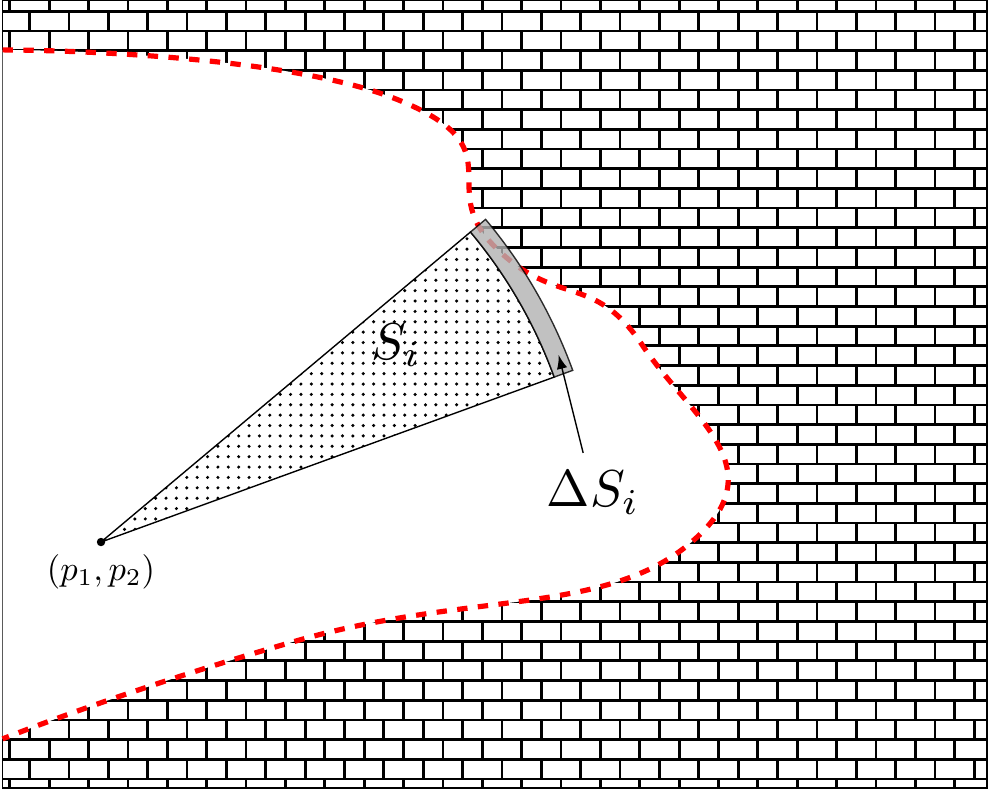}
\par\end{centering}

\label{fig:sensorSets}}

\caption{Sensor model used by the robot.}
\end{figure}

Each sensor emits an acoustic wave in its direction $\alpha_{i}$
which propagates inside a cone of half angle $\gamma$ corresponding
to the aperture of the beam. By measuring the time lag between the
emission and the reception of the wave, reflected by the map, an interval
$[d_{i}]=[d_{i}^{-},d_{i}^{+}]$ contains the true distance $d_{i}$
to the nearest obstacle which lies in the scope of the sensor can
be obtained. The situation is depicted in Figure \ref{fig:sensorModel}.
The area swept by the wave between $0$ and $d_{i}$ is free of obstacles
whereas the map is hit by the wave at distance $d_{i}$. Define
\[
\begin{array}{ccc}
\mathbb{S}_{i} & = & \{(x,y)\mid x^{2}+y^{2}<d_{i}^{-}\text{ and }atan2(y,x)\in[\alpha_{i}-\gamma,\alpha_{i}+\gamma]\}\\
\Delta\mathbb{S}_{i} & = & \{(x,y)\mid x^{2}+y^{2}\in[d_{i}]\text{ and }atan2(y,x)\in[\alpha_{i}-\gamma,\alpha_{i}+\gamma]\}
\end{array}
\]
The set $\mathbb{S}_{i}$ is called the \emph{free} \emph{sector}
and $\Delta\mathbb{S}_{i}$ is called the \emph{impact pie}. These
sets are depicted on Figure \ref{fig:sensorSets}.

The set of all feasible positions $\mathbb{P}$ consistent with $[d_{i}]$
is
\begin{align}
\mathbb{P}(i) & =\{\mathbf{p}\in\mathbb{R}^{2}\mid(\mathbf{p}+\mathbb{S}_{i})\subset\mathbb{M}\text{ and }(\mathbf{p}+\Delta\mathbb{S}_{i})\cap\overline{\mathbb{M}}\neq\emptyset\}\\
 & =(\mathbb{M}\ominus\mathbb{S}_{i})\cap(\mathbb{\overline{M}}\oplus-\Delta\mathbb{S}_{i}).\nonumber 
\end{align}
With several measurements $[d_{i}]$ the set of all positions consistent
with all data is
\begin{align}
\mathbb{P} & =\bigcap_{i}(\mathbb{M}\ominus\mathbb{S}_{i})\cap(\mathbb{\overline{M}}\oplus-\Delta\mathbb{S}_{i}).
\end{align}
Denote by $\mathcal{S}_{\mathbb{M}}$,$\mathcal{S}_{\mathbb{S}_{i}}$,$\mathcal{S}_{\Delta\mathbb{S}_{i}}$
separators for $\mathbb{M}$,$\mathbb{S}_{i}$ ,$\Delta\mathbb{S}_{i}$.
Then a separator for $\mathbb{P}$ is
\begin{equation}
\mathcal{S}_{\mathbb{P}}=\bigcap_{i}(\mathcal{S}_{\mathbb{M}}\ominus\mathcal{S}_{\mathbb{S}_{i}})\cap(\overline{\mathcal{S}_{\mathbb{M}}}\oplus-\mathcal{S}_{\Delta\mathbb{S}_{i}}).
\end{equation}

\begin{figure}[H]
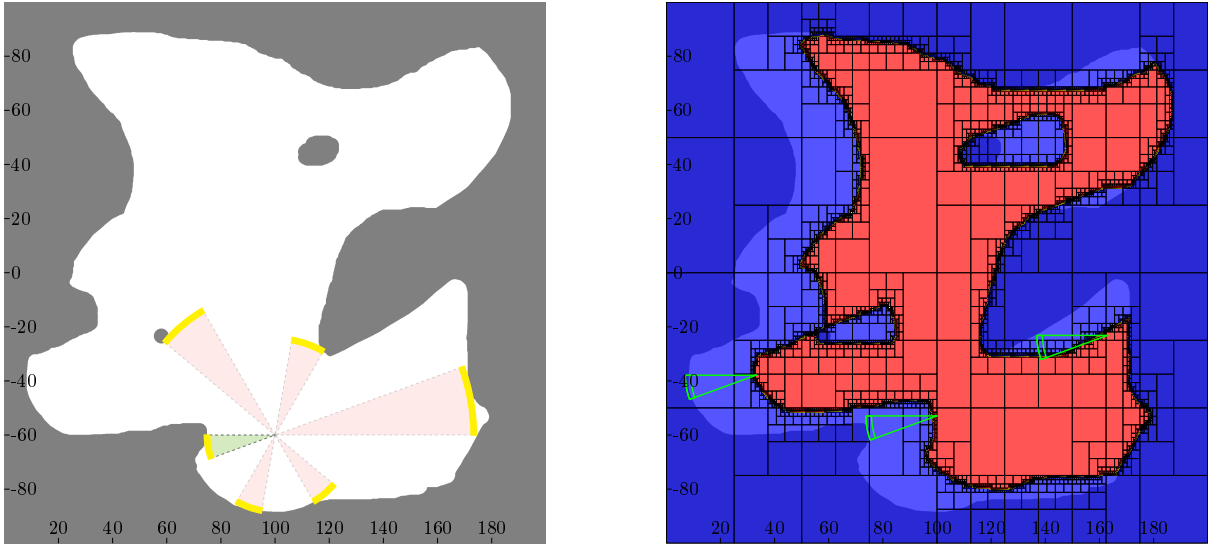

\includegraphics[width=0.45\linewidth]{mink_truth}\hspace*{\fill}\includegraphics[width=0.45\linewidth]{mink_C1_1data}

\caption{Left: a robot which collects 6 sonar range measurements. All free
sectors $\mathbb{S}_{i}$ are included in the map $\mathbb{M}$ (in
white) and the impact pies $\Delta\mathbb{S}_{i}$, in yellow intersects
$\mathbb{\overline{M}}$. Right: Set of all positions for the robot
consistent with the fact that the free sector $\mathbb{S}_{1}$ is
inside $\mathbb{M}$. }
\label{fig:truth:1data}
\end{figure}

As an illustration, consider the situation described by Figure \ref{fig:truth:1data}
(left), where a robot collects 6 sonar data. The width of the intervals
corresponding to the range measurement is $\pm1m$ .The first measurement
corresponding to $i=1$ is painted green. Figure \ref{fig:truth:1data}
(right) corresponds to an approximation of the set $\mathbb{M}\ominus\mathbb{S}_{1}$,
obtained using a paver with the separator $\mathcal{S}_{\mathbb{M}}\ominus\mathcal{S}_{\mathbb{S}_{1}}$. 

\begin{figure}[H]
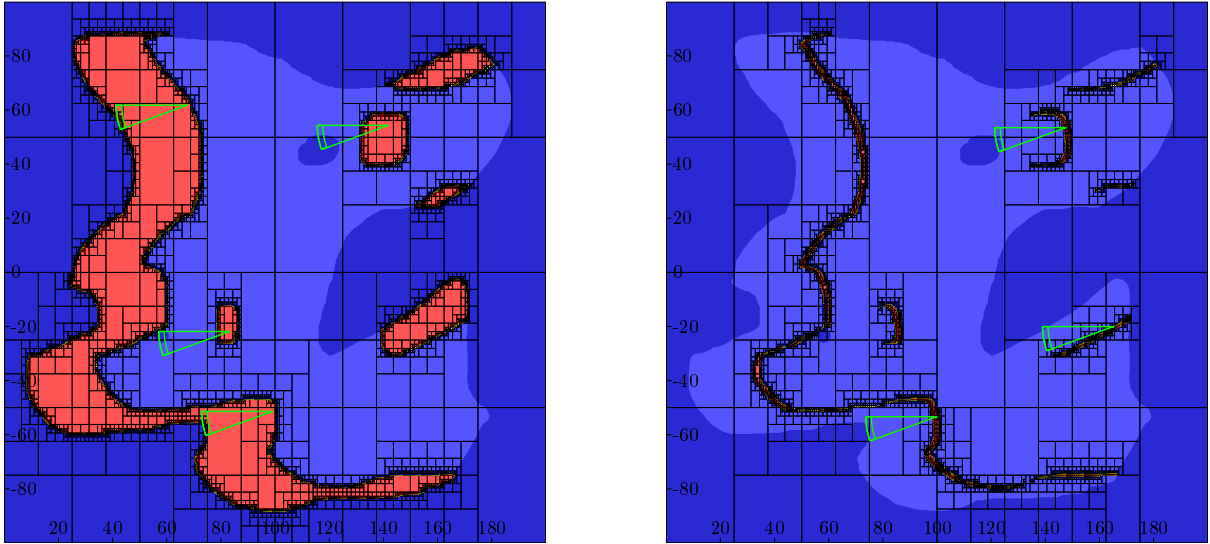

\includegraphics[width=0.45\linewidth]{mink_C2_1data}\hspace*{\fill}\includegraphics[width=0.45\linewidth]{mink_C1C2_1data}

\caption{Left: Positions for the robot consistent with the impact pie $\Delta\mathbb{S}_{1}$.
Right: Positions consistent with the free sector $\mathbb{S}_{1}$
and the impact pie $\Delta\mathbb{S}_{1}$. }
\label{fig:C1:C1C2}
\end{figure}

Figure \ref{fig:C1:C1C2} (left) corresponds to an approximation of
the set $\mathbb{\overline{M}}\oplus-\Delta\mathbb{S}_{1}$, obtained
using a paver with the separator $\overline{\mathcal{S}_{\mathbb{M}}}\oplus-\mathcal{S}_{\Delta\mathbb{S}_{1}}$.
It corresponds to the set of positions for the robot such that the
impact pie $\Delta\mathbb{S}_{1}$ intersects the outside of the map
$\mathbb{M}$. Figure \ref{fig:C1:C1C2} (right) corresponds to the
set $(\mathbb{M}\ominus\mathbb{S}_{1})\cap(\mathbb{\overline{M}}\oplus-\Delta\mathbb{S}_{1}),$
\emph{i.e.}, it contains the position consistent with both $\Delta\mathbb{S}_{1}$
and $\mathbb{S}_{1}$.

\begin{figure}[H]
\includegraphics[width=0.45\linewidth]{mink_C1C2_alldata}\hspace*{\fill}\includegraphics[width=0.45\linewidth]{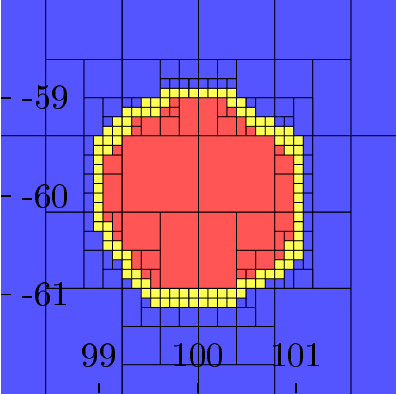}

\caption{Left: Set of positions $\mathbb{P}$ for the robot consistent with
all six free sectors $\mathbb{S}_{i}$ and all impact pies $\Delta\mathbb{S}_{i}$.
Right: zoom around the solution set $\mathbb{P}$.}
\label{fig:all:contractors}
\end{figure}

Figure \ref{fig:all:contractors} (left) corresponds to an approximation
of the set $\mathbb{P}$, obtained using a paver on with the separator
$\mathcal{S}_{\mathbb{P}}$. It corresponds to the set of positions
for the robot that all six impact pies $\Delta\mathbb{S}_{i}$ intersect
the outside of the map $\mathbb{M}$ and all six free sectors are
inside $\mathbb{M}$. A zoom of the solution set is given in Figure
\ref{fig:all:contractors} (right). The computing time is 127 sec.
and 205 boxes have been generated. Note that obtaining an inner approximation
of the solution set was not possible using existing approaches that
are not based on separators.

\section{Conclusion\label{sec:Conclusion}}

Separator-based techniques are particularly attractive when solving
engineering applications, due to the fact that they can handle and
propagate uncertainties in a context where the equations of the problem
are non-linear and non-convex. Now, the performances of paving methods
are extremely sensitive to the accuracy of the separators but also
by the uncertainty generated by the dependency effect induced by the
separator algebra. Indeed, when a separator, associated to the same
set, occurs several times in the separator expression, a pessimism
is introduced. It is thus important to factorize subexpression with
separators into a single one which is computed separately by a specific
algorithm. Another possibility is to rewrite the set expression in
order to avoid multioccurences. This is what we have done for the
Minkowski sum $\mathbb{A}\oplus\mathbb{B}$ and difference $\mathbb{A}\ominus\mathbb{B}$. 

The efficiency of these new operators and their ability to get an
inner and outer approximation of the solution set was illustrated
on the problem of the localization of a robot.

\bibliographystyle{eptcs}
\bibliography{extracted}

\end{document}